\documentclass{article}

\usepackage{microtype}
\usepackage{graphicx}
\usepackage{subcaption}
\usepackage{wrapfig}  
\usepackage{booktabs} 
\usepackage{multirow} 

\usepackage{hyperref}

\usepackage[preprint]{icml2026}

\usepackage{amsmath}
\usepackage{amssymb}
\usepackage{mathtools}
\usepackage{amsthm}

\usepackage[capitalize]{cleveref}

\usepackage{listings}
\theoremstyle{plain}

\theoremstyle{definition}

\theoremstyle{remark}

\usepackage[textsize=tiny]{todonotes}

\newcommand{\method}{Test-time Recursive Thinking\xspace}
\newcommand{\methodshort}{TRT\xspace}

\usepackage{xspace}      
  

\icmltitlerunning{Test-time Recursive Thinking}

\begin{document}

\twocolumn[
  \icmltitle{Test-time Recursive Thinking:\\
  Self-Improvement without External Feedback}



  \icmlsetsymbol{equal}{*}

  \begin{icmlauthorlist}
    \icmlauthor{Yufan Zhuang}{msr,ucsd}
    \icmlauthor{Chandan Singh}{msr}
    \icmlauthor{Liyuan Liu}{msr}
    \icmlauthor{Yelong Shen}{msr}
    \icmlauthor{Dinghuai Zhang}{msr}
    \icmlauthor{Jingbo Shang}{ucsd}
    \icmlauthor{Jianfeng Gao}{msr}
    \icmlauthor{Weizhu Chen}{msr}
  \end{icmlauthorlist}

  \icmlaffiliation{ucsd}{UC San Diego}
  \icmlaffiliation{msr}{Microsoft Research}

  \icmlcorrespondingauthor{Yufan Zhuang}{y5zhuang@ucsd.edu}

  \icmlkeywords{Test-time scaling, ICML}

  \vskip 0.3in
]



\printAffiliationsAndNotice{}  

\begin{abstract}
    Modern Large Language Models (LLMs) have shown rapid improvements in reasoning capabilities, driven largely by reinforcement learning (RL) with verifiable rewards.
    Here, we ask whether these LLMs can self-improve without the need for additional training. 
    We identify two core challenges for such systems: (i) efficiently generating diverse, high-quality candidate solutions, and (ii) reliably selecting correct answers in the absence of ground-truth supervision. 
    To address these challenges, we propose \textit{\method (\methodshort)}, an iterative self-improvement framework that conditions generation on rollout-specific strategies, accumulated knowledge, and self-generated verification signals. Using \methodshort, open-source models reach 100\% accuracy on AIME-25/24, and on LiveCodeBench's most difficult problems, closed-source models improve by 10.4–14.8 percentage points without external feedback. 
    Code is available at \href{https://github.com/EvanZhuang/test_time_recursive_thinking}{link}.
\end{abstract}

\begin{figure}[t]
    \centering
    \includegraphics[width=0.8\columnwidth]{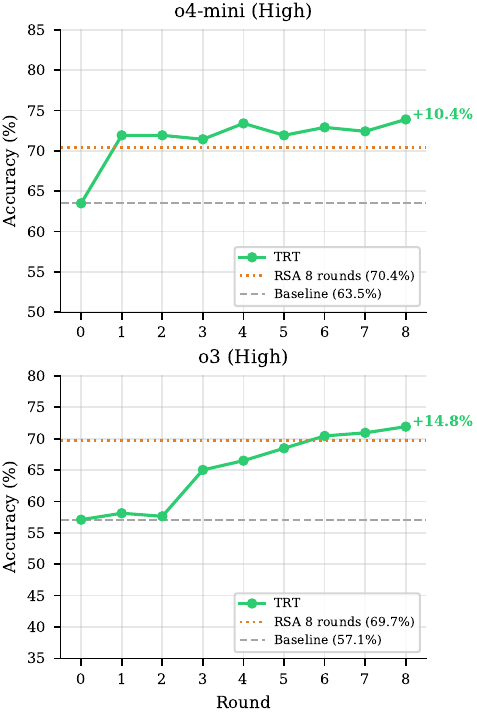}
    \caption{\textbf{LiveCodeBench results.} \methodshort{} accuracy over 8 rounds with test execution on LiveCodeBench v6 hard problems, compared to RSA with 8 rounds. o4-mini improves from 63.5\% to 73.9\% (+10.4 pp), exceeding RSA's 70.4\%; o3 improves from 57.1\% to 71.9\% (+14.8 pp), exceeding RSA's 69.7\%.}
    \label{fig:test_exec}
    \vspace{-2px}
\end{figure}

\begin{figure*}[t]
    \centering
    \includegraphics[width=\linewidth]{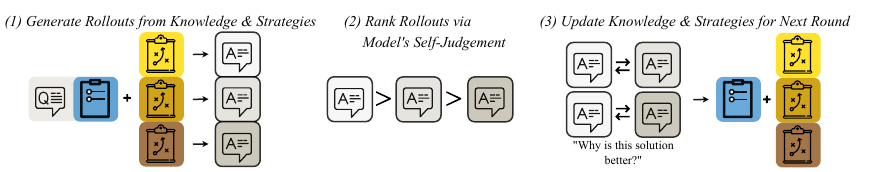}
    \caption{\textbf{\method} iterates between three stages:
    (1) the model generates a set of rollouts conditioned on the current knowledge list and rollout-specific strategies;
    (2) it then self-ranks these rollouts and selects the best solution for the current round;
    (3) finally, the model performs a pairwise analysis comparing the remaining solutions against the best one, distills what to avoid into an updated knowledge list, and synthesizes new rollout strategies to explore in the next round.}
    \label{fig:main_pipeline}
\end{figure*}
\section{Introduction}
The reasoning capabilities of Large Language Models (LLMs) have improved substantially, driven largely by reinforcement learning (RL) with verifiable rewards~\citep{jaech2024openai,guo2025deepseek, shao2024deepseekmath}. However, these paradigms typically depend on external supervision. A critical question remains: can models self-improve their reasoning at test time without access to ground-truth rewards?

Existing approaches fall into two categories.
The first is Meta-RL methods~\citep{duan2016rl,wang2016learning}, which train a model to explore and reflect as a meta-learning objective. 
\citet{qu2025mrt} show that this optimization yields more efficient inference, while \citet{jiang2025lamer} demonstrate that cross-episode training induces exploration through in-context policy adaptation.
While effective, these methods require costly weight updates and complex reward calibration.

A second approach explores self-improvement at inference-time: \citet{zhang2025ace} accumulate and refine strategies through reflection, \citet{venkatraman2025rsa} bootstrap from partially correct steps across reasoning chains,
and \citet{aghajohari2025markovian} handle long reasoning traces with Markovian thinking, maintaining constant-size states across chunks. However, these methods often lack a recursive mechanism to carry learned improvements forward across multiple attempts effectively.

We find that effective test-time self-improvement requires solving two complementary challenges: (1) \textit{strategic exploration} to expand the solution space, and (2) \textit{self-guided verification} to select candidates without ground truth. Neither suffices alone: exploration without verification yields noise, while verification without exploration leads to stagnation. By distinguishing strong solutions from weak ones, a model can extract actionable failure modes and reuse this knowledge to guide subsequent attempts.

We formalize this framework as \method (\methodshort). Given a single problem, the model iteratively: (1) generates multiple rollouts conditioned on accumulated knowledge and exploration strategies, (2) ranks these rollouts using self-judgment, and (3) synthesizes reusable insights by contrasting the best solution against alternatives. These insights are preserved in context, actively guiding future exploration to avoid repeated failures.

We conduct our experiments across mathematical reasoning and code generation. On AIME-25/24, open-source models equipped with \method can achieve 100\% accuracy for the first time.
On LiveCodeBench v6's most difficult problems, \method improves accuracy by 10.4--14.8 pp for o4-mini (high) and o3 (high) respectively.
Together, these results show that LLMs can self-improve at test time by recursively learning from their own attempts, without relying on external feedback.

\section{Related Work}

\paragraph{Meta-Reinforcement Learning and Test-Time Adaptation.}
Meta-RL teaches agents to rapidly adapt to new tasks by balancing exploration and exploitation across episodes~\citep{duan2016rl,wang2016learning}. Recent work has connected meta-RL to test-time compute optimization in LLMs. \citet{snell2024scaling} demonstrates that scaling inference-time computation through search against verifiers or iterative refinement can outperform larger models on reasoning tasks. \citet{qu2025mrt} formalizes test-time compute optimization as a meta-RL problem, showing that optimizing cumulative regret yields 2-3$\times$ relative gains over outcome-reward RL. \citet{jiang2025lamer} applies cross-episode training to LLM agents, showing that meta-RL induces exploration and in-context policy adaptation. These approaches require ground-truth rewards during inference, a requirement our approach eliminates.

\paragraph{Context Engineering and In-Context Self-Improvement.}
A parallel research direction explores test-time self-improvement without weight updates. Reflexion~\citep{shinn2023reflexion} introduces verbal reinforcement learning, where agents reflect on task feedback and maintain reflective text in episodic memory to improve subsequent trials. Self-Refine~\citep{madaan2023selfrefine} demonstrates that LLMs can iteratively generate feedback on their own outputs and refine accordingly, with $\sim$20\% improvements across diverse tasks.
\citet{zhang2025ace} propose Agentic Context Engineering, which treats contexts as evolving playbooks that accumulate and refine strategies through rounds of generation, reflection, and curation. \citet{aghajohari2025markovian} addresses the challenge of ever-growing Chain-of-Thought by proposing Markovian thinking, which decouples thinking length from context size by maintaining constant-size states across reasoning chunks.
These methods either rely on external feedback signals or focus on single-pass refinement rather than parallel rollouts.

\paragraph{Parallel Scaling and Solution Aggregation.}
Parallel test-time scaling generates multiple reasoning paths and aggregates them to identify correct solutions. Self-consistency~\citep{wang2022self} samples diverse reasoning chains and selects answers through majority voting, with gains on arithmetic and commonsense reasoning benchmarks, though recent work on the Sequential Edge~\citep{sharma2025sequential} suggests that inverse-entropy voting in a sequential setting can outperform parallel self-consistency at matched compute. Tree of Thoughts~\citep{yao2023tree} extends chain-of-thought for deliberate exploration over coherent units of text with lookahead and backtracking. Best-of-N sampling with reward models~\citep{cobbe2021training,lightman2023let} uses outcome or process reward models to select the highest-scoring solution from multiple candidates.
Recursive strategies have also emerged to improve efficiency: Recursive self-aggregation (RSA) iteratively refines populations of candidate reasoning chains by aggregating subsets~\cite{venkatraman2025rsa}, while MatryoshkaThinking~\citep{chen2025matryoshkathinking} employs recursive test-time scaling to enable efficient reasoning.
However, these approaches primarily aggregate final answers or merge solutions without extracting \emph{transferable knowledge} about why certain reasoning paths succeed. We address this by distilling actionable insights from contrastive analysis to improve exploration quality, while self-verification enables effective selection without trained reward models.


\section{Methods: \method}
\label{sec:methodology}

Parallel sampling methods generate independent reasoning traces, each unaware of insights from other attempts~\cite{wang2022self, venkatraman2025rsa}. Consequently, models often repeat mistakes and fail to build on partial successes.
To address this, \method (\methodshort) enables knowledge accumulation across iterations and explicit strategy design.
\methodshort operates in $T$ rounds, each consisting of three stages, as shown in \cref{fig:main_pipeline}:

\paragraph{(1) Generate.} Given problem $P$, the model generates $K$ rollouts conditioned on accumulated knowledge $\mathcal{K}$ and rollout-specific strategies $\{s_1, \ldots, s_K\}$:
\begin{equation}
    r_k = \text{LLM}(P, \mathcal{K}, s_k), \quad k \in \{1, \ldots, K\}
\end{equation}
Knowledge constrains the search space, while strategies guide exploration. This produces complementary rather than redundant rollouts.

\paragraph{(2) Select.} The model evaluates the $K$ rollouts and identifies the best solution $r^*$ without ground truth. Selection relies on self-assessment---checking consistency, identifying logical errors, or generating verification tests.

\paragraph{(3) Reflect.} The model compares each non-selected rollout against $r^*$, extracting failure insights that are later appended to the knowledge list
\begin{equation}
    \mathcal{K}_{t+1} = \mathcal{K}_t \cup \{\text{insights from round } t\}
\end{equation}
The model then synthesizes new strategies.

\begin{algorithm}[t]
\caption{\method}
\label{alg:srt}
\begin{algorithmic}[1]
\REQUIRE Problem $P$, rounds $T$, rollouts per round $K$
\ENSURE Selected solution $r^*$
\STATE Initialize knowledge list $\mathcal{K} \gets \emptyset$
\STATE Initialize solution pool $\mathcal{S} \gets \emptyset$
\FOR{$t = 1$ \textbf{to} $T$}
    \STATE \textcolor{gray}{\textit{// Generate}}
    \FOR{$k = 1$ \textbf{to} $K$}
        \STATE Design strategy $s_k$ based on $\mathcal{K}$
        \STATE $r_k \gets \text{LLM}(P, \mathcal{K}, s_k)$
        \STATE $\mathcal{S} \gets \mathcal{S} \cup \{r_k\}$
    \ENDFOR
    \STATE \textcolor{gray}{\textit{// Select}}
    \STATE $r^* \gets \textsc{Select}(\mathcal{S})$
    \STATE \textcolor{gray}{\textit{// Reflect}}
    \FOR{each $r$ in current round where $r \neq r^*$}
        \STATE Extract insights by comparing $r$ to $r^*$
        \STATE $\mathcal{K} \gets \mathcal{K} \cup \{\text{insights}\}$
    \ENDFOR
\ENDFOR\\
\STATE \textbf{return} $r^*$
\end{algorithmic}
\end{algorithm}

\subsection{Core Mechanisms}

\paragraph{Knowledge Representation.}
The knowledge list $\mathcal{K}$ captures domain-specific failure modes (e.g., bug patterns, edge cases, logical fallacies). Entries are phrased as negative constraints (``don'ts") to restrict known bad paths without over-fitting to specific solution steps. To manage context window limitations, the model may prune up to one outdated knowledge entry per round, keeping the knowledge list compact.

\paragraph{Strategy Design.}
To ensure the $K$ rollouts explore distinct regions of the solution space, each rollout receives a unique strategy prompt generated by the model itself. At each round, the model analyzes the accumulated knowledge $\mathcal{K}$ and designs $K$ complementary strategies that avoid previously failed approaches while exploring new directions.

For code generation, strategies specify algorithmic paradigms (e.g., \textit{dynamic programming} vs. \textit{greedy} vs. \textit{divide and conquer}), implementation priorities (\textit{optimize for memory} vs. \textit{optimize for speed}), or structural choices (\textit{iterative} vs. \textit{recursive}). For mathematical reasoning, strategies may emphasize different proof techniques (\textit{algebraic manipulation} vs. \textit{geometric intuition}) or problem decomposition approaches (\textit{work backwards} vs. \textit{case analysis}).

The model generates strategies conditioned on knowledge of what has failed, enabling targeted exploration rather than redundant attempts. This self-directed strategy design distinguishes \methodshort from methods that rely on fixed strategy pools or random variation.

\begin{figure*}[t]
    \centering
    \includegraphics[width=\textwidth]{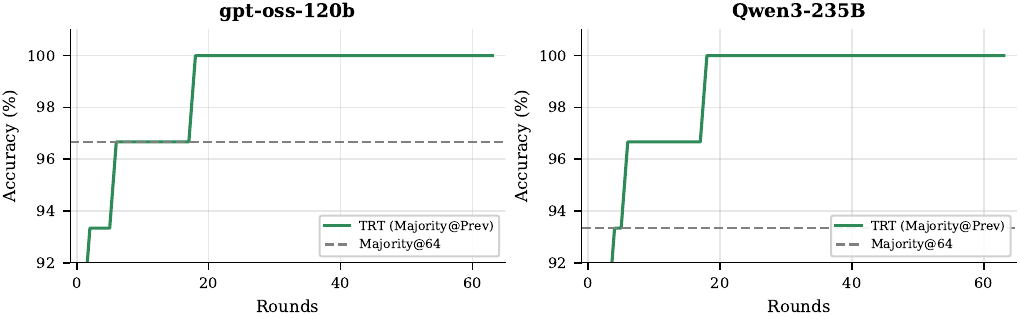}
    \caption{\textbf{AIME-25 results.} \method achieves 100\% accuracy for both gpt-oss-120b and Qwen3-235B. The rolling majority vote (Majority@Prev) shows monotonic improvement, outperforming the Parallel Thinking baseline (Majority@64).}
    \label{fig:aime_main}
\end{figure*}

\begin{figure*}[t]
    \centering
    \includegraphics[width=\textwidth]{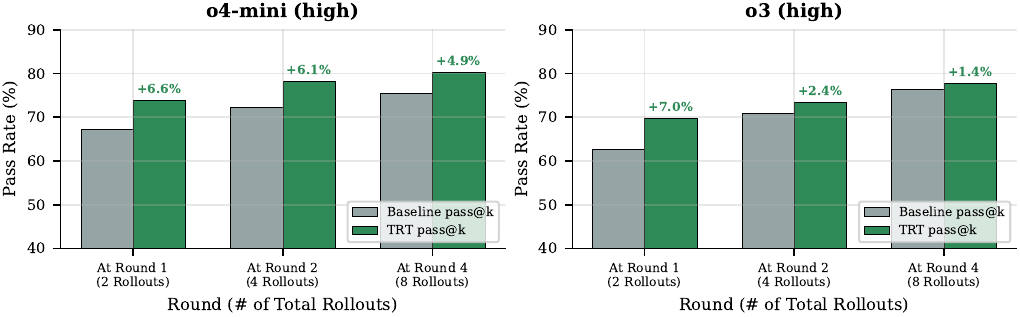}
    \caption{\textbf{Rollout exploration efficiency comparison.} At equivalent sample counts, \method's strategic planning consistently outperforms independent sampling by 2-7 pp in pass@k across both models.}
    \label{fig:rollout_analysis}
\end{figure*}

\subsection{Domain Specific Designs}
\label{sec:domain_instantiations}

Since \methodshort lacks access to ground truth, the selection mechanism ($\textsc{Select}$) must exploit the specific structural properties of the domain.

\paragraph{Mathematical Reasoning: Mutual Exclusivity.}
For problems with a single correct integer answer (e.g., AIME), we exploit the property of \textit{mutual exclusivity}. The answer space $\mathcal{A}$ contains exactly one correct answer $a^*$. While incorrect answers disperse across the remaining space, correct reasoners converge on $a^*$. We track all the previously self-rejected answers in the knowledge list, to help model self-assess the correctness of their generated solution at each round. 

\paragraph{Code Generation: Execution-Based Self-Verification.}
In coding, multiple valid implementations exist, rendering mutual exclusivity ineffective. However, code allows for empirical verification. We employ a test-based selection mechanism where the model generates unit tests derived from its understanding of the problem.

Concretely, for each candidate solution, the model generates a suite of test cases covering typical inputs, edge cases, and boundary conditions based on its interpretation of the problem specification. Each candidate is then executed against these tests. Solutions that pass more tests are ranked higher. When multiple solutions pass all generated tests, the model applies secondary criteria: preferring solutions with cleaner logic, better handling of edge cases mentioned in the problem, or consistency with accumulated knowledge about what approaches have previously failed.

This self-verification is imperfect: model-generated tests may miss edge cases that ground-truth tests would catch. However, it provides a meaningful selection signal without external feedback. The ablation in~\cref{tab:ablation} shows that test execution contributes 7.4 percentage points of improvement, confirming its effectiveness for discriminating between candidates.

The knowledge list for code tracks bug patterns, edge cases that caused failures, and algorithmic insights. Unlike math where knowledge focuses on solution approach, code knowledge emphasizes implementation pitfalls: off-by-one errors, performance bottlenecks, and format mismatches that caused test failures in previous rounds (see detailed analysis in~\cref{fig:knowledge_categories}).

\section{Main Results}


\label{sec:experiments}

We evaluate \methodshort on AIME (mathematical reasoning) and LiveCodeBench (code generation). Our experiments address two questions: (1) Can models self-improve via iterative knowledge accumulation? (2) How does \method compare to parallel sampling at equivalent compute?

\subsection{Experimental Setup}

\paragraph{Datasets.}
For mathematical reasoning, we use the 30 problems from AIME 2025~\cite{aime} as our primary mathematics benchmark. These competition-level problems require multi-step reasoning and have unambiguous integer answers (0-999). We also conducted experiments on the 30 problems from AIME 2024~\cite{aime2024} following the same methodology, results are included in~\cref{app:aime24}.
For code generation, we evaluate on 203 hard problems from LiveCodeBench v6~\cite{jain2024livecodebench}. 

We consider math experiments as preliminary due to their relatively small dataset size.
They serve as a proof-of-concept.
We design and implement deeper analysis for code generation where the dataset is larger and convenient mutual exclusiveness and majority voting are no longer available to the model. 

\begin{figure*}[t]
    \centering
    \includegraphics[width=\textwidth]{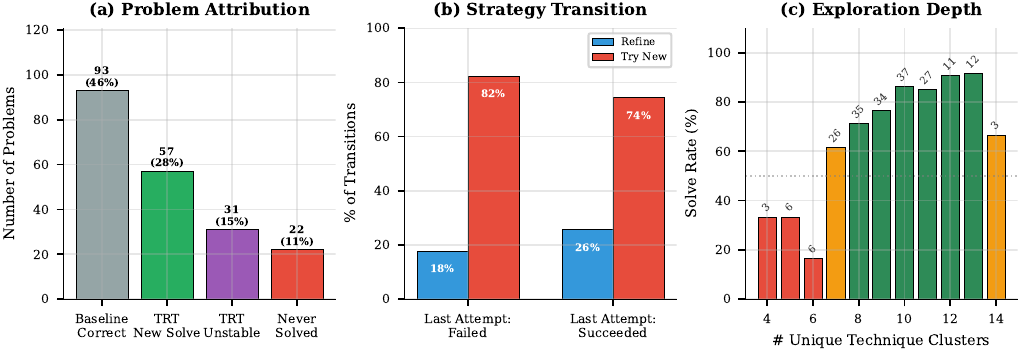}
    \caption{\textbf{Problem-level and strategy analysis} (o4-mini with test execution). (a) Problem attribution by round 8 state: \textit{Baseline Correct} (baseline solved, TRT retained), \textit{TRT New Solve} (baseline failed, TRT solved), \textit{TRT Unstable} (baseline solved, TRT lost), \textit{Never Solved} (both failed). Categories sum to Selected accuracy (73.9\%). (b) Strategy transitions: models switch more frequently after failure (82\%) than success (74\%). (c) Exploration depth: solve rate increases with the number of unique technique clusters explored.}
    \label{fig:analysis_combined}
\end{figure*}

\paragraph{Models.}
For AIME, we rely on large open-weights models, \textit{gpt-oss-120b} and \textit{Qwen3-235B-A22B-Thinking}~\cite{agarwal2025gpt, yang2025qwen3}, to incorporate their transparent reasoning traces in knowledge generation. Conversely, for LiveCodeBench, we utilize the proprietary \textit{o3 (High)} and \textit{o4-mini (High)}~\cite{openai_o3_o4mini_system_card_2025}, as they provide superior performance on competitive programming and executable code generation.

\paragraph{Baselines.}
We compare against \textit{Parallel Thinking with Majority Vote}, which generates multiple independent reasoning traces and aggregates answers via majority voting, and \textit{Recursive Self-Aggregation (RSA)}~\citep{venkatraman2025rsa}, which iteratively refines populations of candidate solutions by aggregating subsets' outputs. 

We ensure fair comparison by using equivalent compute budgets. On AIME, we compare 64 independent traces with majority voting against 64 refinement rounds of \method ($K{=}1$) with majority aggregation over rounds. On LiveCodeBench, we compare \method with 2 rollouts per round over 8 rounds against RSA with population 2 with 8 iterations. 

The choice of $K=\{1,2\}$ in our main experiments is justified with our ablations with various $K$ (see~\cref{fig:rollout_scaling}); we found that increasing $K$ does not lead to clear improvements on the performance upper bound when total round $T$ is large enough. 

\paragraph{Test Execution.}
For code generation, the model also generates its own test cases based on problem understanding and uses execution results to guide selection, providing self-generated feedback without ground-truth access.

\paragraph{Metrics.}
For AIME, we report accuracy. For LiveCodeBench, we report \textbf{accuracy} (accuracy of the selected solution of each round), \textbf{Cumulative Best} (best accuracy achievable with oracle across all rounds) and \textbf{pass@k} (pass rate with accumulated $k=K\times t$ rollouts at round $t$). The gap between Cumulative Best and per-round accuracy of the selected solution represents recoverable performance through improved selection, whereas the pass@k indicates how much exploration the model has done with the rollout budget.
    
\subsection{Preliminary Results: Mathematical Reasoning}

\cref{fig:aime_main} presents our AIME-25 results. We evaluate using 64 refinement rounds ($K{=}1$) with rolling majority vote aggregation across rounds, comparing against 64 independent traces with majority voting.

\method achieves 100\% accuracy on all 30 problems with both models, solving every problem through iterative knowledge accumulation. Crucially, both models also stabilize around 100\% at the \textit{instance level} near the end of 64 rounds and remain stable across multiple runs (\cref{fig:stability}). This instance-level result holds across AIME 2024 as well (\cref{fig:aime24}).

The performance improves monotonically across rounds, confirming that accumulated knowledge guides exploration towards the right direction.

\subsection{Main Results: Code Generation}

Code generation is substantially harder than mathematical reasoning on AIME. Unlike math problems with a single correct integer answer, programming problems admit many valid implementations and require the model to reason about algorithmic correctness, edge cases, and execution behavior. Moreover, majority voting over answers is no longer applicable, making effective exploration and selection significantly more challenging.

\cref{fig:test_exec} reports performance over 8 rounds of iterative refinement. Both models exhibit consistent and monotonic improvement, demonstrating that \methodshort{} enables sustained self-improvement in open-ended code generation.

For o4-mini, accuracy improves from 63.5\% in the first round to 73.9\% after 8 rounds, a gain of +10.4 percentage points. This exceeds the performance of Recursive Self-Aggregation (RSA), which achieves 70.4\%, by 3.5 points. Similarly, o3 improves from 57.1\% to 71.9\% (+14.8 percentage points), surpassing RSA’s 69.7\% by 2.2 points.

\subsection{Validations of the Key Designs}

\paragraph{Does \methodshort improve rollout exploration efficiency?}
\label{sec:rollout_analysis}

We first examine whether strategic planning per roll-out and knowledge accumulation improved the system's coverage in the solution space. On LiveCodeBench with 2 rollouts per round ($K=2$), we compare pass@$k$ against baseline (sampling randomly with temperature=1) at equivalent sample counts. 

Strategic planning improves pass@k by 2--7 percentage points across both models (\cref{fig:rollout_analysis}), indicating that accumulated knowledge and per-rollout strategy indeed improved the model's exploration in the solution space.

\begin{figure}[t]
    \centering
    \includegraphics[width=0.8\columnwidth]{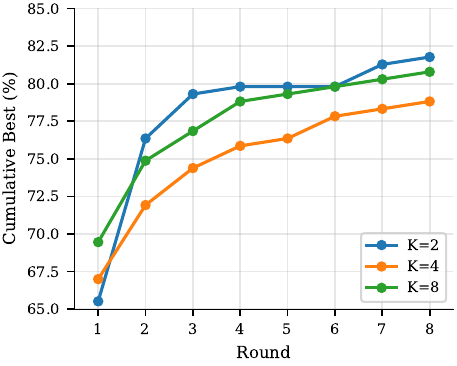}
    \caption{\textbf{Effect of number of iterations versus number of rollouts (o4-mini).} Cumulative best accuracy over rounds for different rollouts per round ($K \in \{2, 4, 8\}$). All three configurations converge to similar upper bounds, demonstrating that performance gains come primarily from iterative knowledge accumulation rather than parallel sampling breadth.}
    \label{fig:rollout_scaling}
\end{figure}

\paragraph{Effect of Depth $T$ and Width $K$}
We examine whether increasing rollouts per round ($K$) improves the achievable upper bound beyond $K{=}2$ (\cref{fig:rollout_scaling}). For o4-mini, $K{=}2$, $K{=}4$, and $K{=}8$ all converge to similar cumulative best accuracy (78--82\%). If more rollouts cannot raise the ceiling, there is no benefit to parallel breadth within rounds. This confirms that depth beats breadth, iterative refinement across rounds matters more than parallel exploration within rounds.

\paragraph{Ranking \& Selection Needs Grounding}

\begin{table}[h]
\centering
\small
\caption{\textbf{LiveCodeBench ablation.} Self-generated test execution reduces the selection gap significantly (from 11.8\% to 4.9\% for o4-mini, 16.3\% to 3.5\% for o3).}
\resizebox{\columnwidth}{!}{%
\begin{tabular}{llccc}
\toprule
Model & Configuration & Acc. & Cumul. Best & Gap \\
\midrule
\multirow{3}{*}{o4-mini}
  & Baseline (pass@1) & 63.5\% & --- & --- \\
  & + Strategy & 66.5\% & 78.3\% & 11.8\% \\
  & + Strategy + Test Exec & \textbf{73.9\%} & 78.8\% & \textbf{4.9\%} \\
\midrule
\multirow{3}{*}{o3}
  & Baseline (pass@1) & 57.1\% & --- & --- \\
  & + Strategy & 61.6\% & 77.8\% & 16.3\% \\
  & + Strategy + Test Exec & \textbf{71.9\%} & 75.4\% & \textbf{3.5\%} \\
\bottomrule
\end{tabular}%
}
\label{tab:ablation}
\vspace{-4px}
\end{table}

We further decompose the gains to determine whether they arise from exploration (per-rollout strategy \& knowledge list) or verification (self-designed test execution). Table~\ref{tab:ablation} shows both contribute substantially. From baseline 63.5\%, strategic planning adds 3.0 percentage points by discovering solutions unreachable through random sampling. Self-generated test execution contributes 7.4 percentage points through effective selection. 

These results indicate that \methodshort{} improves code generation through two complementary mechanisms. First, accumulated knowledge and per-rollout strategy guide exploration toward more promising solutions, leading to higher pass@k compared to parallel sampling with the same rollout budget. Second, self-generated tests and execution feedback enable the model to better discriminate between candidate programs, narrowing the gap between cumulative best and the selected solution in each round. Together, these effects allow \methodshort{} to outperform RSA under equivalent compute, while avoiding the need for ground-truth labels or external feedback.

\section{Analysis}

To understand \methodshort better, we analyze with the following questions: how compactly insights are stored, what the contribution of \methodshort is, and how exploration adapts across rounds.

\subsection{Growth of Knowledge List Length}
\label{sec:knowledge_efficiency}

\begin{figure}[t]
    \centering
    \includegraphics[width=0.8\columnwidth]{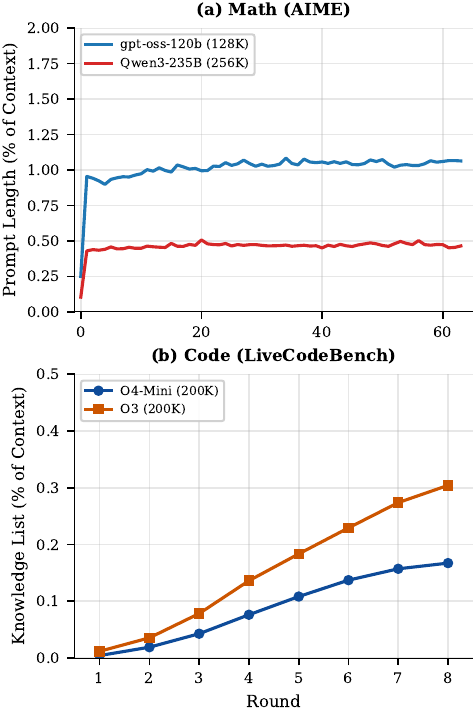}
    \caption{\textbf{Knowledge list length as percentage of maximum context.} The accumulated knowledge list remains compact across both domains. \textbf{(a)} For math tasks, the list never exceeds 1.5\% after 64 rounds: gpt-oss-120b averages 1.00\% (max 1.07\%) of its 128K context, while Qwen3-235B averages 0.47\% (max 0.52\%) of its 256K context. \textbf{(b)} For code tasks, the list stays below 0.35\% after 8 rounds: o4-mini averages 0.16\% (max 0.31\%) and o3 averages 0.09\% (max 0.17\%) of their 200K contexts. This efficiency enables \methodshort to scale without context exhaustion.}
    \label{fig:knowledge_list_length}
\end{figure}

Effective context management is a critical part in iterative reasoning frameworks, where the accumulation of lengthy knowledge often leads to context window saturation and subsequent performance degradation.

Figure~\ref{fig:knowledge_list_length} illustrates the token efficiency of \methodshort, demonstrating that the method maintains minimal memory overhead across diverse domains.
For math tasks (AIME), the list never exceeds 1.5\% of context even after 64 rounds. For code tasks (LiveCodeBench), efficiency is even higher, staying below 0.35\% after 8 rounds. 

This high compression ratio is a direct result of our decision to persist only distilled, high-level insights rather than retaining verbose solution trajectories or raw chain-of-thought data. By prioritizing the storage of abstracted error corrections over full interaction histories, \methodshort enables sustainable scalability for long-horizon tasks without inducing context exhaustion.

\subsection{Problem-Level Breakdown}
\label{sec:problem_analysis}

In \cref{fig:analysis_combined}a, we categorize the 203 problems by baseline accuracy and \methodshort outcomes. \textit{Baseline Correct} (46\%) were solved by baseline and retained by \methodshort. \textit{TRT New Solve} (28\%) were unsolved by baseline but solved by \methodshort at round 8. \textit{TRT Unstable} (15\%) were solved by baseline but lost by \methodshort during refinement. \textit{Never Solved} (11\%) remained unsolved by both approaches. 

Of particular interest are 36 problems (17.7\%) that we term \textit{breakthroughs}: problems where baseline failed even at pass@10, yet \methodshort eventually solved. These problems are unreachable through independent sampling within practical sample budgets. Strategic exploration finds qualitatively different solutions than random sampling. Appendix~\ref{app:case_studies} presents detailed case studies of one such breakthrough problem, illustrating how accumulated knowledge guides strategy evolution.

\subsection{Strategy Dynamics}

We analyze how models adapt their strategies across rounds using SoftTFIDF~\cite{cohen2003comparison} on the per-rollout strategy text to measure strategy similarity between consecutive rounds (\cref{fig:analysis_combined}b).

Models adapt their exploration: strategy switches occur more frequently after failure (solution failed ground truth tests, note the model never sees the actual outcome) (82\%) than success (solution passed ground truth tests) (74\%), showing that the model can sense its progress and its exploration strategy accordingly.

We further show that there is a positive correlation between the number of attempted strategies, and the end-of-trajectory performance (\cref{fig:analysis_combined}c): problems with more unique technique clusters explored show higher solve rates, strategic diversity improves outcomes.

\subsection{Knowledge Accumulation Patterns}
\begin{figure}[t]
    \centering
    \includegraphics[width=0.9\columnwidth]{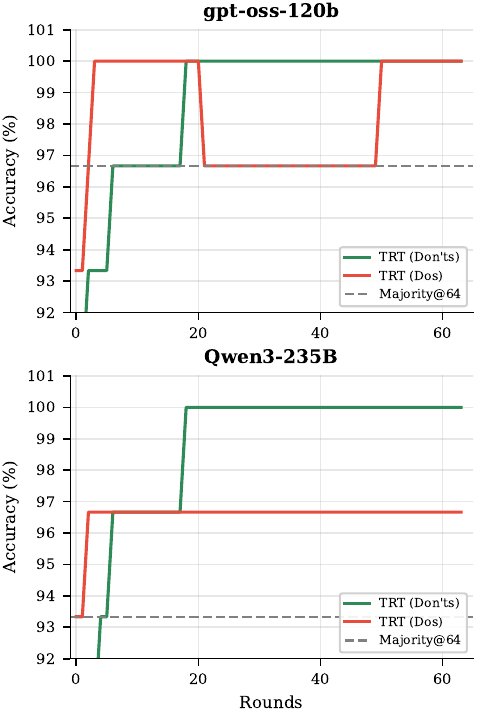}
    \caption{\textbf{Knowledge type comparison on AIME-25.} Recording what not to do (Don'ts) yields higher accuracy than recording successes (Dos) for both models.}
    \label{fig:knowledge_type}
\end{figure}

\begin{figure}[t]
    \centering
    \includegraphics[width=0.8\columnwidth]{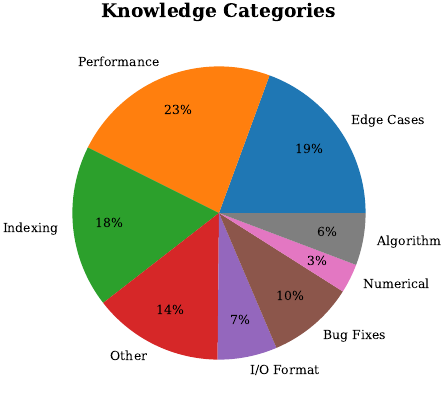}
    \caption{\textbf{Knowledge categories in code generation.} Performance optimizations and edge cases dominate, reflecting common failure modes in competitive programming.}
    \label{fig:knowledge_categories}
\end{figure}

On AIME-25, recording failures yields stronger performance gains than reinforcing successes (\cref{fig:knowledge_type}). Based on this insight, we design the knowledge system for both math and code.

We grouped the accumulated knowledge from LiveCodeBench into categories via keyword matching, as shown in~\cref{fig:knowledge_categories}.
The distribution of accumulated knowledge reveals that improvements are driven primarily by higher-level execution insights rather than low-level syntactic corrections. As shown in \cref{fig:knowledge_categories}, the largest portions of learned knowledge concern \textit{Performance} (23\%), \textit{Edge Cases} (19\%), and \textit{Indexing} (18\%), indicating that the model most frequently records lessons about efficiency, boundary conditions, and off-by-one or access patterns. In contrast, categories such as \textit{Bug Fixes} (10\%) and \textit{I/O Format} (7\%) account for a smaller share, while purely \textit{Algorithmic} (6\%) and \textit{Numerical} (3\%) knowledge is relatively rare.

\section{Discussion}

\paragraph{Summary.}
\method shows that LLMs can self-improve within a single problem instance by iterating on their own rollouts, accumulating knowledge along the way.
Our experiments confirm the central hypothesis: it is important to both encourage efficient exploration via knowledge accumulation \& per-rollout strategy design, and to have effective self-guided solution selection mechanisms.

\paragraph{Limitations.}
While effective, \methodshort demonstrates some limitations.
Domains like coding may require specific adaptations (e.g., test execution environments).
Test-generation quality also varies by problem type, and the residual selection gap indicates room for better verification.
Computational overhead scales linearly with rounds: $T{=}8$ rounds with $K{=}2$ rollouts requires 16 samples plus 8 reflection calls, compared to 16 samples for pass@16. This additional cost is offset by improved sample efficiency.

\paragraph{Future Directions.}
Future work could aggregate knowledge across problems rather than single instances to improve generalization on an entire class of problems~\cite{qu2025rlad,agrawal2025gepa}.
Since the aggregated knowledge is stored as text (or human-readable code), the knowledge itself could be of interest in scientific discovery tasks~\cite{reddy2025towards,wang2025thetaevolve,singh2023explaining}.
An alternative line of work may continue to improve \methodshort capabilities, e.g. by using reinforcement learning to improve recursive thinking at training time, as some works have done for different variations of reasoning~\cite{zhan2025exgrpo,aghajohari2025markovian,jiang2025meta}.
Alternatively, \methodshort could be tailored to particular domains, e.g. by using verification mechanisms that extend to mathematical proof checking,
or using human supervision to tackle problems requiring external knowledge~\cite{wen2025scalable,feng2026human}.

\section*{Impact Statement}
This paper presents work whose goal is to advance the performance of LLMs at inference time, which has many potential societal consequences, including increased computational/energy cost and improved capabilities of LLMs when used for nefarious purposes.


{
    \small
    \bibliography{ref}
    \bibliographystyle{icml2026}
}

\newpage
\appendix
\onecolumn
\section{Appendix}

\counterwithin{figure}{section}
\counterwithin{table}{section}
\renewcommand{\thefigure}{\thesection\arabic{figure}}
\renewcommand{\thetable}{\thesection\arabic{table}}

\subsection{Breakthrough Case Studies}
\label{app:case_studies}

We present a case study of problems where \methodshort achieved breakthroughs: problems the baseline model failed to solve even with 10 independent attempts, but which \methodshort solved through iterative refinement.

\subsubsection{Problem 192: ``Replace'' (AtCoder ABC 399-E)}
\label{app:case_replace}

\paragraph{Problem.} Given two binary sequences $A$ and $B$ of length $N$ with associated costs $C_i$, minimize the total cost to transform $A$ into $B$ via weighted flips, where each flip costs $C_i$ multiplied by the current number of ones in the sequence.

\paragraph{Trajectory.} \texttt{[X X X X O O O O]} --- Baseline failed at pass@10; \methodshort achieved breakthrough in round 5 and maintained stability through round 8.

\paragraph{Strategy Evolution.}
\begin{itemize}
    \item \textbf{Rounds 1--4}: Initial strategies attempted greedy scheduling (sorting flips by cost) and parametric search. These achieved high internal test pass rates but failed on ground truth due to overlooking the role of ``neutral'' bits (positions where $A_i = B_i = 1$) that can be temporarily flipped to reduce intermediate sums.
    \item \textbf{Round 5}: The model switched to a \emph{Two-Pointer Sweep} approach that explicitly tracks neutral flips. By sorting neutrals in descending order and maintaining running prefix sums, this strategy evaluates the cost function for each count $k$ of neutral flips in $O(1)$ per step, finding the true minimum.
\end{itemize}

\paragraph{Accumulated Knowledge.}
By round 4, the model had accumulated the following insights that informed the successful round-5 strategy:
\begin{itemize}
    \item \emph{``The naive block-greedy approach failed by rigidly performing all $1{\to}0$ flips before $0{\to}1$ flips and never using flips of neutral bits to reduce the running sum. Optimal schedules interleave neutral flips.''}
    \item \emph{``Rigid greedy ordering fails to capture cost reductions achievable by interleaving neutral flips. One must parameterize on the number of neutral flips and use prefix sums to evaluate the total cost function.''}
    \item \emph{``Use bisect\_right for strict greater than x counts and bisect\_left for strict less than x to avoid off-by-one errors at boundaries.''}
\end{itemize}

\subsection{Ablation: Sequential Edits vs Rewriting}
\label{app:edit_vs_regen}

\begin{figure}[h!]
    \centering
    \includegraphics[width=\columnwidth]{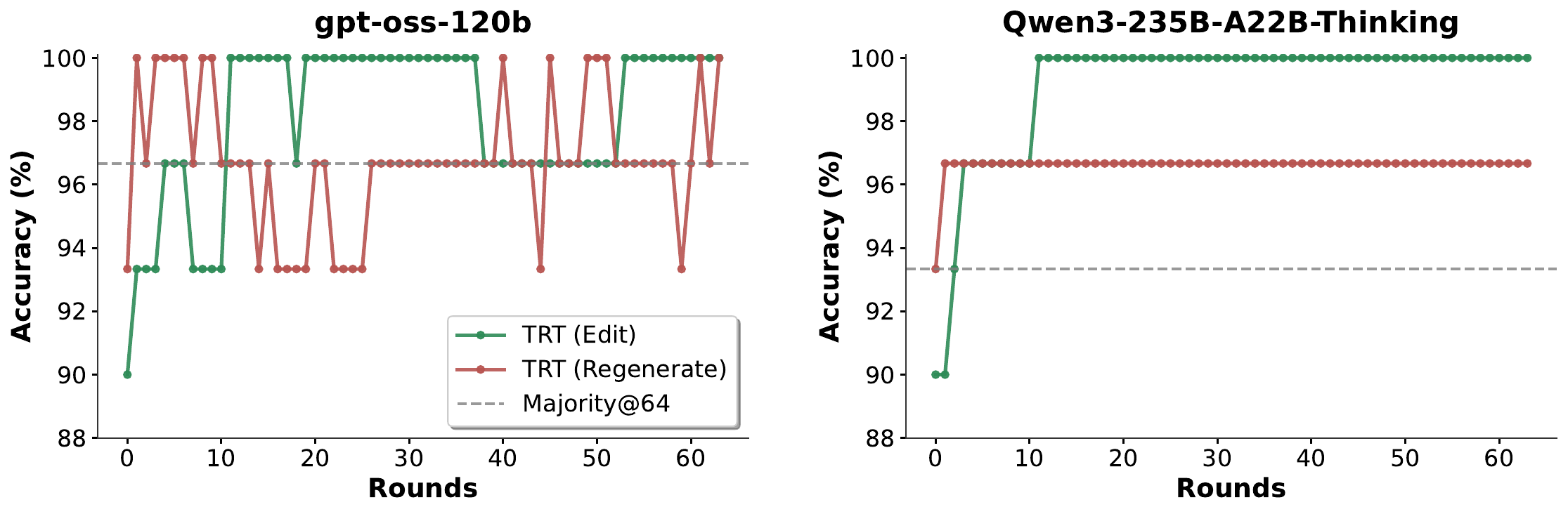}
    \caption{\textbf{Sequential edits outperform full regeneration.} When the model incrementally edits its previous solution based on accumulated knowledge (Edit), it achieves higher accuracy than when generating solutions from scratch each round (Regenerate). This suggests that preserving working components while fixing identified issues is more effective than starting fresh.}
    \label{fig:edit_vs_regen}
\end{figure}

We compare two solution generation strategies: (1) sequential editing, where the model refines its previous solution based on new knowledge, and (2) full regeneration, where the model generates a new solution from scratch each round using only the accumulated knowledge list. Figure~\ref{fig:edit_vs_regen} shows that sequential editing consistently outperforms regeneration. This advantage likely stems from the ability to preserve working components (e.g., correct parsing logic) while targeting identified failure modes.

\subsection{Small Model Results}
\label{app:small_models}

\begin{figure*}[t]
    \centering
    \includegraphics[width=\textwidth]{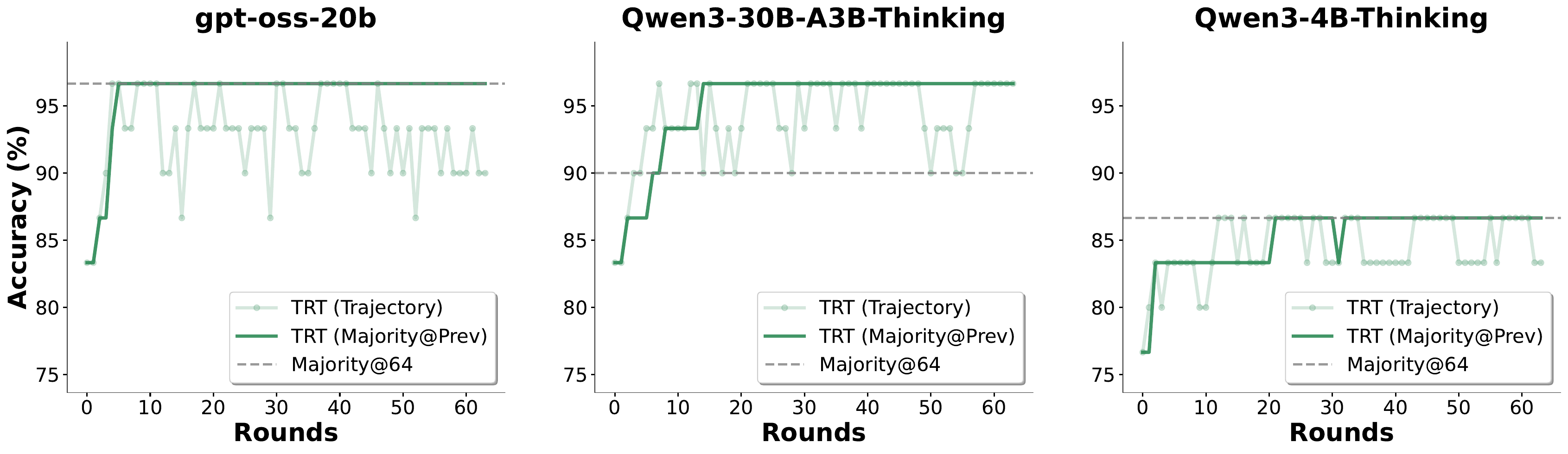}
    \caption{\textbf{\methodshort on smaller models (AIME-25).} Left to right: gpt-oss-20B, Qwen3-30B, and Qwen3-4B. All models benefit from iterative knowledge accumulation, with majority voting over previous rounds (solid line) consistently outperforming individual round accuracy (faded line). Notably, even the 4B parameter Qwen3 model achieves competitive performance, suggesting \methodshort's benefits extend across model scales.}
    \label{fig:small_models}
\end{figure*}

To assess whether \methodshort's benefits extend to smaller models, we evaluate on three models: gpt-oss-20B, Qwen3-30B, and Qwen3-4B (Figure~\ref{fig:small_models}). All models show improvement through iterative refinement, with majority voting over accumulated rounds consistently outperforming both individual round accuracy and the Majority@64 baseline. The baseline majority@64 accuracies are 96.7\% (gpt-oss-20B), 90.0\% (Qwen3-30B), and 86.7\% (Qwen3-4B), and \methodshort achieves competitive or superior performance on all three.

\subsection{AIME-24 Results}
\label{app:aime24}

\begin{figure*}[t]
    \centering
    \includegraphics[width=0.9\textwidth]{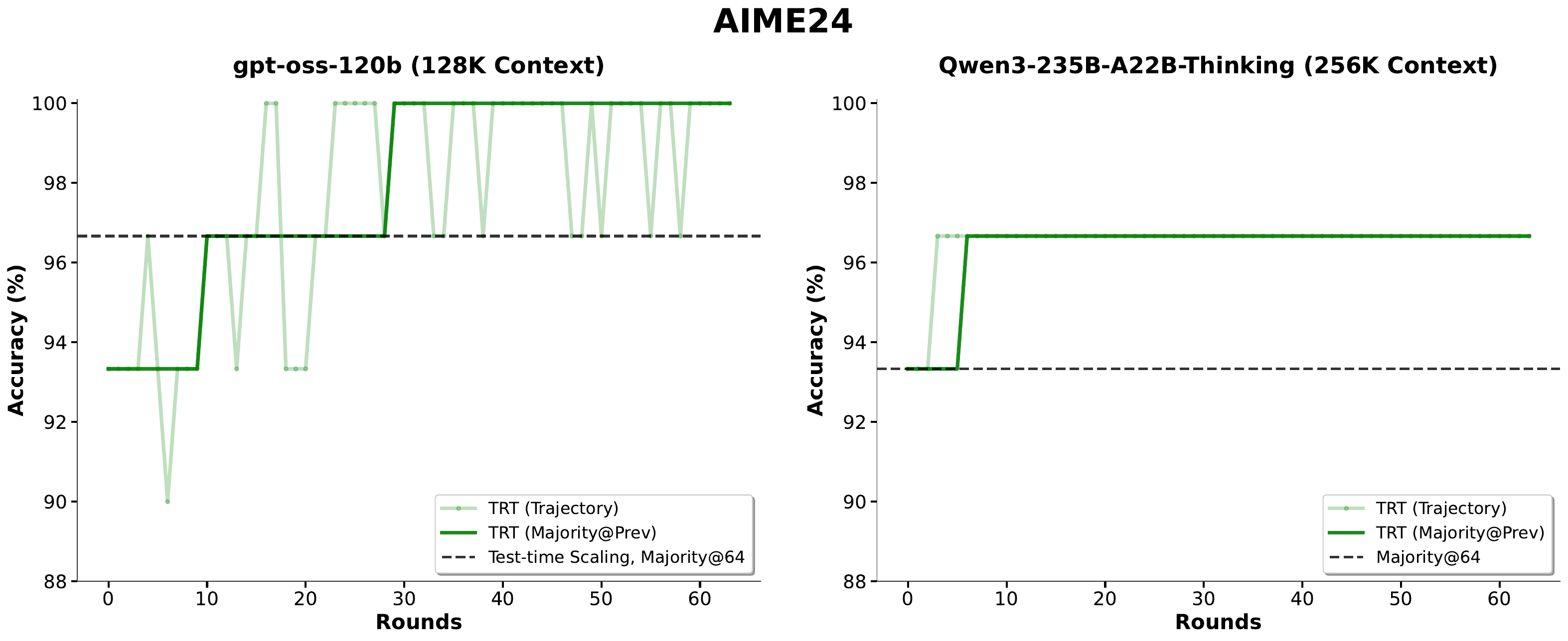}
    \caption{\textbf{AIME-24 results.} \methodshort generalizes to the AIME-24 benchmark, showing similar improvement patterns as observed on AIME-25. Both gpt-oss-120b and Qwen3-235B demonstrate consistent gains through iterative knowledge accumulation.}
    \label{fig:aime24}
\end{figure*}

To verify that \methodshort's improvements are not specific to AIME-25, we evaluate on the AIME-24 benchmark (Figure~\ref{fig:aime24}). The results show similar improvement trajectories, confirming that the iterative knowledge accumulation mechanism generalizes across different problem sets within the same domain.

\subsection{AIME Stability}
\label{app:reproducibility}

\begin{figure}[t]
    \centering
    \includegraphics[width=\columnwidth]{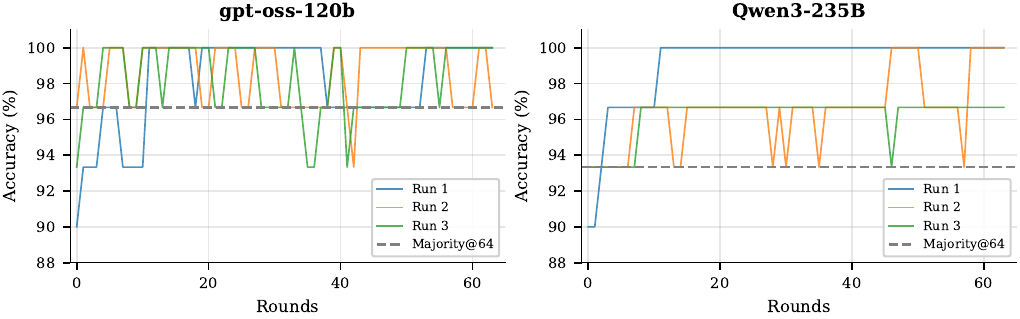}
    \caption{\textbf{Reproducibility on AIME-25.} Three independent runs show consistent performance across both models. gpt-oss-120b achieves 100\%, 96.7\%, and 100\% accuracy across runs. Qwen3-235B achieves 100\%, 100\%, and 96.7\%. All runs outperform their respective Majority@64 baselines.}
    \label{fig:stability}
\end{figure}

\begin{figure*}[t]
    \centering
    \includegraphics[width=\textwidth]{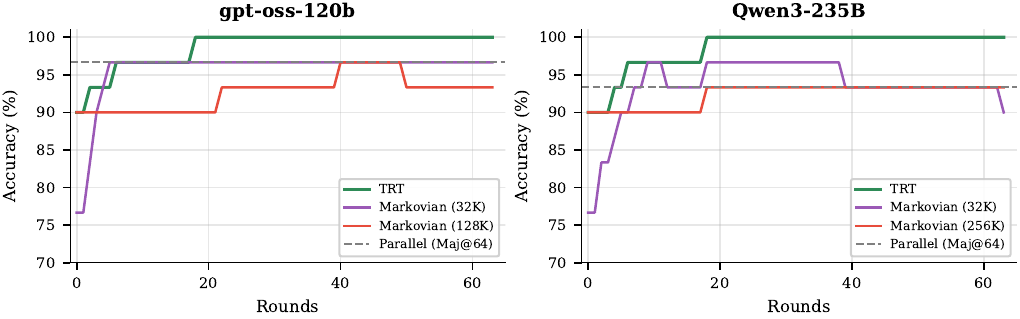}
    \caption{\textbf{Comparison with Markovian Thinking on AIME-25.} \methodshort{} (green) achieves monotonic improvement through structured knowledge accumulation. Markovian Thinking~\citep{aghajohari2025markovian} appends previous answers to extend context: at limited 32K context (purple), it recovers near-baseline performance; at full context (red), accuracy degrades as accumulated answers dilute the reasoning signal.}
    \label{fig:markovian_comparison}
\end{figure*}

To assess reproducibility, we conduct three independent runs for each model on AIME-25 (Figure~\ref{fig:stability}). All runs outperform their respective Majority@64 baselines (96.7\% and 93.3\%), demonstrating that \method produces reliable improvements with low variance.

\subsection{Comparison with Markovian Thinking}
\label{app:markovian}

An alternative approach to multi-round reasoning is Markovian Thinking~\citep{aghajohari2025markovian}, which appends previous answers to the context without structured knowledge extraction. \cref{fig:markovian_comparison} compares these approaches.

At limited context (32K tokens), Markovian Thinking recovers near-baseline performance, suggesting the iterative paradigm itself is sound. However, at full context (128K--256K tokens), performance plateaus at approximately 93\% as accumulated CoT dilutes the reasoning signal. \methodshort{} avoids this by extracting compressed knowledge---what to avoid, what worked---occupying less than 1.5\% of context while providing targeted guidance (see~\cref{fig:knowledge_list_length}). This enables stable performance improvement beyond the context limit.

\subsection{System Prompts}
\label{app:prompts}

We provide the core system prompts used in \methodshort. These prompts define the behavior of the solver agent, knowledge manager, and test generation components.

\begin{lstlisting}[title={\textbf{Solver System Prompt}}]
You are an expert competitive programmer solving coding problems.

## YOUR TASK:
1. Read the problem statement carefully
2. If you have previous attempts/solutions shown, analyze them critically for bugs
3. Write a complete, working Python solution
4. If a reference solution exists, analyze it critically for bugs. Your job is to improve it.

## CRITICAL OUTPUT FORMAT (MANDATORY):
Your solution MUST be wrapped in a Python code block like this:
```python
# your code here
```

## SOLUTION QUALITY:
- Ensure your solution solves the problem as stated
- Check time/space complexity against problem constraints
- Test your logic mentally with the given examples

## VERIFICATION:
Before finalizing, trace through your solution with the example inputs from the problem.
If your output doesn't match the expected output, you have a bug - fix it before submitting.
\end{lstlisting}

\begin{lstlisting}[title={\textbf{Knowledge Manager System Prompt}}]
You are a technical reviewer analyzing competitive programming solutions.

Your job:
1. Rank solutions by CORRECTNESS first, then EFFICIENCY
2. Extract SPECIFIC, ACTIONABLE lessons from failures (not generic tips)
3. DEDUPLICATE insights - don't repeat what's already in knowledge base
4. Update the knowledge base with the best solution

DO NOT solve the problem yourself. Focus only on evaluation and knowledge curation.
IMPORTANT: Call update_knowledge exactly ONCE at the end.
\end{lstlisting}

\begin{lstlisting}[title={\textbf{Test Generator Prompt}}]
You are a test engineer designing test cases for competitive programming.

Your task:
1. Analyze the problem and candidate solutions
2. Generate discriminating test cases that differentiate solutions
3. Call execute_generated_tests EXACTLY ONCE

Prioritize tests that:
- Target differences in logic between solutions
- Expose bugs: off-by-one, boundary conditions, edge cases
- Include problem examples for baseline validation
\end{lstlisting}

\subsubsection{AIME Prompt Templates}
\label{app:aime_prompts}

For mathematical reasoning tasks (AIME), we use structured prompt templates that guide the model through iterative refinement.

\begin{lstlisting}[title={\textbf{AIME Initial Prompt}}]
Guideline: Let's solve this problem. Be thorough.

## Output format (Use exact headers including square brackets):
[Summary]: A paragraph of detailed step-by-step summary of your solution, write
thoroughly and in details, note down every steps of calculation you did, and
what was the final answer you got.
[Answer]: Therefore, final answer is \boxed{<integer>}.

Let's think step by step. Follow the output format strictly.
\end{lstlisting}

\begin{lstlisting}[title={\textbf{AIME Iterative Refinement Prompt}}]
Let's solve this problem. I have some additional information that might help.
Examine them carefully and see if they can help you solve the problem more
accurately.

{knowledge_text}

### Reference Solution
Take these information with a grain of salt, they might be wrong or incomplete.
Try to spot the mistakes in the solution and see if there is a more accurate
approach.
{reference_solution}

### Output format (Use exact headers including square brackets):
[Why the reference solution is wrong?]: If you get a different solution than
the reference solution, explain here in a stand-alone manner, you must explain
what is the reference solution's final answer, and why is it incorrect.
(or write "N/A" if you agree with the reference solution)
[Summary]: A paragraph of detailed step-by-step summary of your solution, write
thoroughly and in details, note down every steps of calculation you did, and
what was the final answer you got.
[Answer]: Therefore, final answer is \boxed{<integer>}.

Let's think step by step. Follow the output format strictly.
\end{lstlisting}

The \texttt{\{knowledge\_text\}} placeholder is populated with an ``Empirical Mistakes List'' containing previous wrong answers and explanations of why they were incorrect. The \texttt{\{reference\_solution\}} placeholder contains the summary from the previous round's solution attempt.






\end{document}